\documentclass[journal]{IEEEtran}

\usepackage{cite}

\ifCLASSINFOpdf
\else
\fi
\usepackage{amsmath}
\usepackage{upgreek}

\ifCLASSOPTIONcompsoc
  \usepackage[caption=false,font=normalsize,labelfont=sf,textfont=sf]{subfig}
\else
  \usepackage[caption=false,font=footnotesize]{subfig}
\fi
\usepackage{url}

\usepackage{breakurl}

\usepackage{pgfplots}
\usepackage{tikz}
\usetikzlibrary{shapes,arrows}
\definecolor{gold}{rgb}{1.0, 0.84, 0.0}
\definecolor{blue citius}{rgb}{0.03529412, 0.29411765, 0.49803922}
\xdefinecolor{green citius}{rgb}{0,0.6,.1}
\definecolor{royalpurple}{rgb}{0.47, 0.32, 0.66}
\definecolor{apricot}{rgb}{0.98, 0.81, 0.69}
\definecolor{processcyan}{rgb}{0.0, 0.72, 0.92}
\definecolor{springgreen}{rgb}{0.0, 1.0, 0.5}
\definecolor{orchid}{rgb}{0.85, 0.44, 0.84}

\begin{document}
%
\title{Deep Learning-Based Multiple Object Visual Tracking on Embedded System for IoT and Mobile Edge Computing Applications}
%
%
%

\author{Beatriz Blanco-Filgueira, 
		Daniel Garc\'ia-Lesta,  
		Mauro Fern\'andez-Sanjurjo,
        V\'ictor Manuel Brea 
        and Paula L\'opez
\thanks{This work has been submitted to the IEEE for possible publication. Copyright may be transferred without notice, after which this version may no longer be accessible. This work has been partially supported by the Spanish government project TEC2015-66878-C3-3-R MINECO (FEDER), the Conseller\'ia de Cultura, Educaci\'on e Ordenaci\'on Universitaria (accreditation 2016-2019, ED431G/08, and reference competitive group 2017-2020, ED431C 2017/69) and the European Regional Development Fund (ERDF).
}        
\thanks{The authors are with the Centro Singular de Investigaci\'on en Tecnolox\'ias da Informaci\'on, Universidade de Santiago de Compostela, R\'ua de Jenaro de la Fuente Dom\'inguez, 15782 Spain.}
}

\maketitle

\begin{abstract}
Compute and memory demands of state-of-the-art deep learning methods are still a shortcoming that must be addressed to make them useful at IoT end-nodes. In particular, recent results depict a hopeful prospect for image processing using Convolutional Neural Netwoks, CNNs, but the gap between software and hardware implementations is already considerable for IoT and mobile edge computing applications due to their high power consumption. This proposal performs low-power and real time deep learning-based multiple object visual tracking implemented on an NVIDIA Jetson TX2 development kit. It includes a camera and wireless connection capability and it is battery powered for mobile and outdoor applications. A collection of representative sequences captured with the on-board camera, dETRUSC video dataset, is used to exemplify the performance of the proposed algorithm and to facilitate benchmarking. The results in terms of power consumption and frame rate demonstrate the feasibility of deep learning algorithms on embedded platforms although more effort to joint algorithm and hardware design of CNNs is needed.
\end{abstract}


%
\IEEEpeerreviewmaketitle

\section{Introduction}
%
%
%
%

\IEEEPARstart{V}{isual} tasks such as object detection, classification or tracking are essential for many practical applications at the perception or sensor layer of the IoT architecture~\cite{Rusci2017}. These features rely on a large quantity of data and require prompt response, that is, the scene must be captured and processed for real time decision making. Defining real time as live performance, understood as the capability of solving the application problem using at any time only the available frame from the live camera without storing intermediate frames for delayed processing, this requirement can not be efficiently accomplished by cloud computing due to latency and the prospect of poor coverage. In such cases computation is moved to the edges of the networks, that is, the sensor layer of IoT, giving rise to what is called edge computing~\cite{Shi2016}. The energy efficiency of such visual sensing nodes is one of the key criteria that lead their design since interconnected devices of the IoT infrastructure need to be battery-powered in many situations, such as mobile and outdoor applications. Even further, they could be self-powered by means of energy harvesting techniques, for instance. 

Image processing solutions range from traditional computer vision algorithms to the more recent application of deep learning-based strategies. The potential of the later has led image processing to a new extent during the last years~\cite{LeCun2015}. Convolutional neural networks, known as CNNs~\cite{Lecun98}, along with recent computational capabilities, offer a new strategy to address computer vision tasks. Among the advantages of CNNs in comparison with traditional computer vision algorithms are their robustness and accuracy. Whereas traditional algorithms are optimized for a concrete goal and particular conditions, CNNs are trained in a massive way to undertake more general challenges. Although the training phase is usually highly demanding in terms of computation, the benefits of the CNN can be exploited using less sophisticated hardware resources once the CNN has been trained.

The results of the last years depict a hopeful prospect for image processing using CNNs. Efforts have been focused on a wide range of applications such as image classification and object detection~\cite{He2016}, segmentation~\cite{Shelhamer2017} and object tracking~\cite{gundogdu2018good}. Despite the promising results in most of them, object tracking using deep features and CNNs has only recently emerged. One of the state-of-the-art benchmarks for object tracking is the Visual Object Tracking (VOT) challenge~\cite{vot} and the winners of the last years were based both on deep learning techniques~\cite{nam2016learning,bertinetto2016fully} and deep features~\cite{danelljan2016beyond,danelljan2017eco}.

Despite the rapid development of deep learning methods, and CNNs for computer vision purposes particularly, the gap between software and hardware implementations is already considerable~\cite{Sze2017}. In order to make state-of-the-art networks useful at IoT end-nodes, more attention must be paid to their power consumption and compute and memory demands. Most of the CNNs energy consumption is related to data movement rather than computation itself~\cite{Keckler2011}. Hence, it is necessary to make a great effort not only at network design but also at hardware implementation in order to exploit CNNs potential in the computer vision field for low-power real time mobile systems and IoT applications. In fact, some authors have recently highlighted the importance of the joint algorithm and hardware design of neural networks~\cite{Sze2017}.

Algorithms and CNNs are evolving and competing continuously to be more accurate and faster and they are usually tested with benchmark databases over powerful hardware platforms. On the other hand, manufacturers and scientists try to diversify the available hardware options to provide the required resources to implement and accelerate the performance of those demanding networks. However, there is a lack of end-to-end IoT end-nodes performing image capture, processing and communication. For this purpose, embedded platforms such as NVIDIA Jetson TX2 are probably the most suitable choice in terms of design speed and cost effectiveness to develop proof of concept and even final solutions to a wide range of applications. Moreover, Jetson TX2 is offered in a ready to use development kit but also as a single board computer with a size of 50x87~mm and 85~g weight. The development kit includes a camera and wireless connection capability, thus development from beginning to end can be accomplished from an early stage~\cite{jetsontx2}.

In this work, low-power and real time deep learning-based multiple object tracking is implemented on an NVIDIA Jetson TX2 development kit. Section~\ref{sec:hardware} offers a discussion about hardware availability and our choice. Next, our own implementation of the GOTURN CNN tracker as a multiple object tracking proposal is described in Section~\ref{sec:algorithm}. A Hardware-Oriented PBAS (HO-PBAS) algorithm~\cite{garcia2017effect} is used to detect moving objects and it is integrated with the GOTURN CNN based tracker~\cite{Held2016}. Moreover, additional code was included to manage multiple object tracking. Section~\ref{sec:results} shows the performance of the proposed algorithm over a collection of representative sequences captured with the on-board camera, dETRUSC video dataset~\cite{citius}, and the results in terms of power consumption and velocity. Finally, the main conclusions are summarized in Section~\ref{sec:conclusions}.

\section{Hardware implementation} \label{sec:hardware}

CNNs offer increasing accuracy for visual tasks at the expense of huge computation and memory resources. Current hardware solutions can satisfy these demands at the cost of high energy consumption, specially when real time processing is also a requirement, understood as the capability of solving the application problem using at any time only the available frame provided by the live camera without intermediate storage for delayed processing. That is the case of edge computing solutions, where the computation is performed near the data source in order to avoid cloud transfer and thus improve response time~\cite{Shi2016}. Edge computing still benefits from energy savings due to no cloud transfer need, but it also requires embedded low-power consumption nodes. In order to exploit the sate-of-the-art CNNs for real applications at IoT end-nodes, low-power embedded solutions must be explored whereas maintaining reasonable accuracy and performance.

Diverse hardware solutions for CNN inference acceleration have been proposed in recent years. They range from standalone solutions to heterogeneous systems and Systems-on-Chip (SoC), which can include field-programmable gate arrays (FPGAs), application-specific integrated circuits (ASICs), CPUs and graphics processing units (GPUs). 

Focusing on visual tasks-oriented proposals, those based on FPGAs stand out in terms of energy efficiency but not in performance~\cite{Nurvitadhi2017,Guo2018}. Additionally, some of them are supported in other elements as CPUs or external DRAM memory~\cite{Gokhale2014,Qiu2016,Shah2018,Ma2018}, the networks used as benchmark are not always representative~\cite{Wang2017,Zhang2015} or their price limits the application range~\cite{Qiao2017,Bettoni2017}.	

In terms of ASICs, several accelerators for CNNs as an alternative to CPUs and GPUs for computer vision tasks as image classification~\cite{Chen2017}, face detection~\cite{Bang2017} and recognition~\cite{Du2017}, have been presented during the last year. They present a very low consumption but long development cycles, high cost and little flexibility to adapt to the rapid progress of the algorithms.

Regarding CPUs, Intel Knights Landing~\cite{KnightsLanding} and Intel Knights Mill~\cite{KnightsMill} generations of Xeon Phi x86 CPUs are optimized for deep learning but conceived for training in supercomputers, servers and high-end workstations. Intel also offers the Aero Compute Board, a purpose-built unmanned aerial vehicle (UAV) developer kit powered by a quad-core Intel Atom processor. 
  
NVIDIA, one of the most important GPU manufacturers, has also increased its offering for deep learning solutions rapidly~\cite{nvidiaAI}. NVIDIA GPUs for deep learning cover data center solutions (DGX systems and Tesla solutions), desktop development (Titan Xp, Quadro GV100, Titan V) and embedded applications (Jetson TX2). NVIDIA DGX systems are fully integrated solutions built on the NVIDIA Volta GPU platform but server oriented. NVIDIA DGX-2 contains 16 Tesla V100 GPUs consuming 10~kW whereas NVIDIA DGX-1 is based on 8 Tesla V100 GPUs with 3.5~kW consumption. Facebook's Big Sur~\cite{bigsur} and more recent Big Basin~\cite{bigbasin} custom deep learning server, with 8 NVIDIA Tesla P100 GPU accelerators, is similar to NVIDIA DGX-1. 

NVIDIA Jetson TX2 is a promising AI SoC powered by NVIDIA Pascal GPU architecture for inference at the edge. It is power-efficient, with small dimensions and high throughput for embedded applications. Whereas discrete GPUs consumption ranges between 150 and 250~W, integrated GPU as on the Jetson TX2 swings between 5 and 15~W. Additional advantages are the reduced size and needless active cooling. It is also commercially available as a development kit~\cite{jetsontx2}, which also includes a 5MP CSI camera and WLAN and Bluetooth connectivity among others peripherals. Since Jetson TX2 development kit fits all requirements, including a reasonable cost, it was chosen for the scope of this work. 

Jetson TX2 consists of a quad-core 2.0~GHz 64~bit ARMv8 A57 processor, a dual-core 2.0~GHz superscalar ARMv8 Denver processor, and an integrated Pascal GPU 1.3~GHz with 256 cores. The six CPU cores and the GPU share 8 GB DRAM memory. Jetson TX2 includes a command line tool for switching operation modes at run time, adjusting the CPUs and GPU clock speeds by Dynamic Voltage and Frequency Scaling (DVFS), see Table~\ref{tx2modes}. Max-Q mode represents the peak of the power/throughput curve, that is, the peak efficiency, which corresponds to 7.5~W consumption and limits the clocks to ensure operation in the most efficient range only. Max-P enables maximum system performance although with higher consumption (15~W maximum) and reduced efficiency. Custom configurations with intermediate frequencies are also allowed for the purpose of balancing between peak efficiency and peak performance. Finally, DVFS can be disabled to run all cores at the maximum speed all time activating full clocks mode while operating at any mode except Max-Q.

\begin{table}[!t]
	\renewcommand{\arraystretch}{1.3}
	\caption{NVIDIA Jetson TX2 Operation Modes}
	\label{tx2modes}
	\centering
	\begin{tabular}{|c||c||c||c|}
	\hline
	Mode & Denver (GHz) & A57 (GHz) & GPU (GHz) \\ \hline
	Max-N & 2.0 & 2.0 & 1.30 \\ \hline
	Max-Q & - & 1.2 & 0.85 \\ \hline
	Max-P Core-All & 1.4 & 1.4 & 1.12 \\ \hline
	Max-P ARM & - & 2.0 & 1.12 \\ \hline
	Max-P Denver & 2.0 & - & 1.12 \\ \hline
	\hline
	\end{tabular}
\end{table}

\begin{figure}[!t] \centering
	\includegraphics[width=\columnwidth]{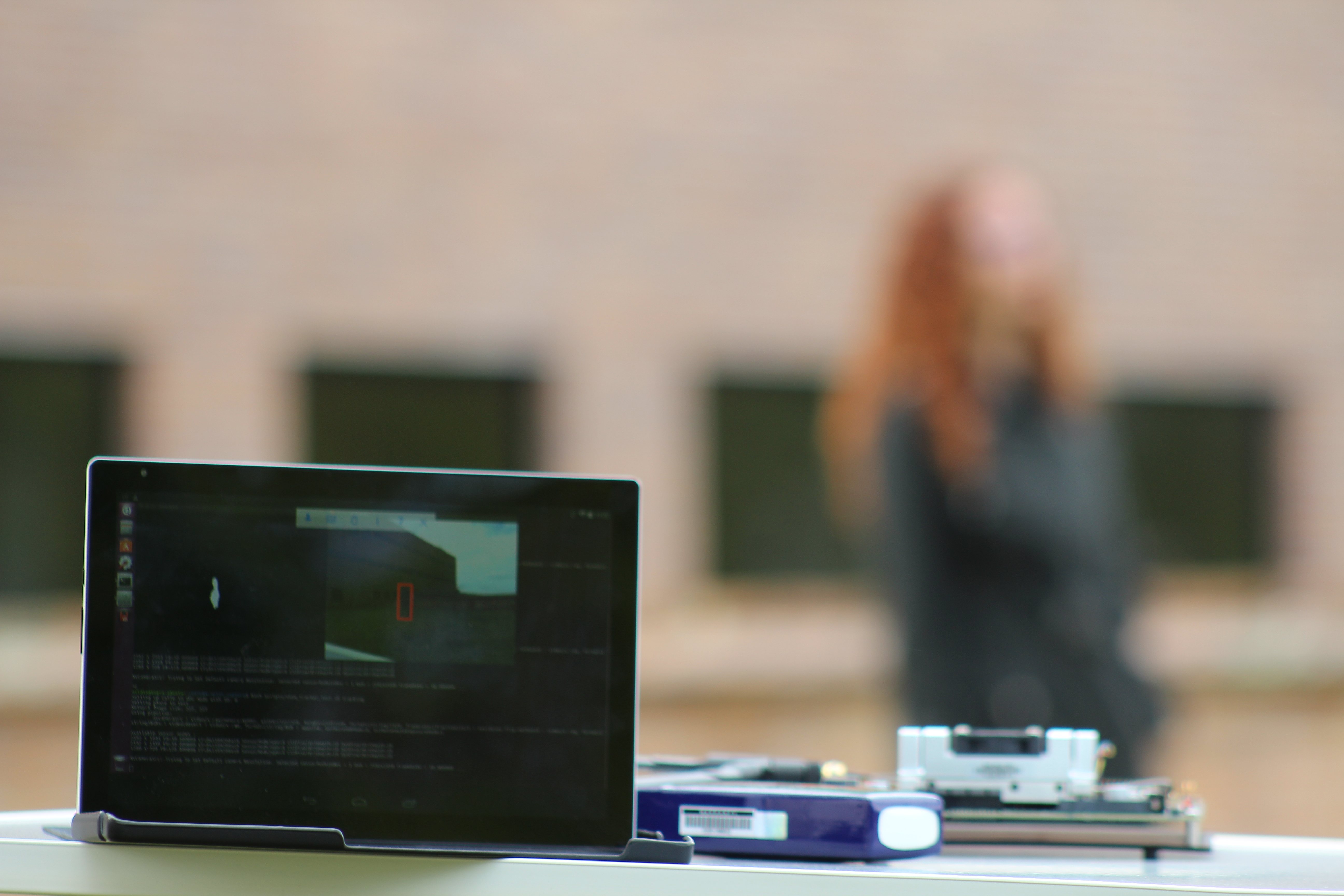}
	\caption{Embedded solution: Jetson TX2 development kit is battery powered and remotely controlled using a tablet and WiFi connection.}
	\label{fig:setup}
\end{figure}

A picture of our experimental set-up during execution is shown in Fig.~\ref{fig:setup}. Jetson TX2 development kit is powered by a 3S LiPo battery and remotely controlled by a tablet using WiFi connection. The images shown on the tablet correspond to live performance, where the black and white image on the left represents the detection of the person in the background and the camera capture with a red box on the right depicts the tracking. More details can be found in Section~\ref{sec:accuracy}.

\section{Multiple object tracking approach} \label{sec:algorithm}

\subsection{Overview}

A proposal for multiple object tracking was developed making use of the Hardware-Oriented PBAS (HO-PBAS) algorithm~\cite{garcia2017effect} to detect moving objects integrated with the GOTURN CNN based tracker~\cite{Held2016}. The HO-PBAS detector~\cite{garcia2017effect} is a hardware oriented foreground segmentation method based on the pixel-based adaptive segmenter (PBAS) algorithm~\cite{hofmann2012background}. It is oriented to focal-plane processing, \cite{suarez2017low}, with the benefit of less memory usage, becoming appropriate for future on-chip implementation of the whole algorithm and also for its use on embedded solutions such as the Jetson platform. On the other side, GOTURN~\cite{Held2016}, that was originally designed for 1-object tracking and implemented using Caffe~\cite{jia2014caffe}, has been applied in this work to multiple object tracking. It was tested on VOT2014 dataset~\cite{vot2014} and it is one of the the few that can achieve real time performance with the additional advantage of reasonable hardware requirements. 

Both algorithms have been tested on publicly available datasets and present good results separately. In this work, they were integrated to address an end-to-end solution over the NVIDIA Jetson TX2 embedded platform described in Section~\ref{sec:hardware}. It consists in multiple object tracking of real time detected objects. For this purpose, both algorithms must run jointly, obtaining precise object detections from HO-PBAS which are used as inputs for the GOTURN tracker. Regardless of the accuracy of the used detector, its output can not be as precise as manual annotation of the object to track, which is the approach commonly used for tracker validation such as in VOT competition. Thus, merely integration of both algorithms is not enough and additional development was needed to overcome limitations of a non ideal input annotation for the tracker. 

A diagram representation of the proposed algorithm is depicted in Fig.~\ref{fig:diagram}. The model and the camera are initialized and then detection and tracking of multiple objects is carried out as it is explained in the following subsections.

\subsection{Detection}

The output of the HO-PBAS detector are the bounding boxes of the moving objects in the scene which have an area between desired maximum and minimum values. As new objects are detected in the scene, their bounding boxes are sent to the tracker. Typically, trackers are tested on annotated videos where the first frame bounding box is used as input, that is, the object to track. Thus, the object is completely at the scene in the first frame and the manual annotation is ideal. This means that the object can be well characterized by the CNN and the tracker algorithms can be compared in the same terms for challenge purpose, regardless of the detection. However, in a real situation objects come into the scene from the edges and they are detected as soon as they occupy a minimum area, what can occur before they enter the image completely. If a partially cropped object is sent to the tracker, it probably does not manage enough information to treat it as the same object when it appears completely. Therefore, partial object detection must be prevented by only considering completely detected objects. In our implementation, this is accomplished by requiring the bounding boxes to be separated from the image borders by a certain number of pixels.

\subsection{Tracking of previous objects and new candidates}

\begin{enumerate}
	\item After the detection, if moving objects were already detected in previous frames (``Previous objects?'', Fig.~\ref{fig:diagram}), the GOTURN algorithm continues tracking them independently of HO-PBAS performance. To do that, the CNN itself estimates the new position of every tracked object. Additionally, in order to stop the tracking of those that leave the scene, the algorithm checks if any of the bounding boxes are close to the edges of the image (``Track (GOTURN) and stop (if needed)'', Fig.~\ref{fig:diagram}).
	\item Then, it must be checked whether there is a new object in the scene. The minimum matching between all pairs of previous and current bounding boxes, calculated as the intersection over union~\cite{WuLimYang13}, is selected as a potential new detection (``Match previous and current detections'', Fig.~\ref{fig:diagram}). In the particular case that more than one new object appeared on the scene, they would be identified and their tracking initialized in subsequent frames.
	\item After that, it must be determined whether the new detection is really a new object (``New object?'', Fig.~\ref{fig:diagram}). This is solved by comparing the new detection with all the bounding boxes previously estimated by the tracker. This prevents false positives derived, for instance, from the following circumstances: 
\end{enumerate}
 
\begin{itemize}
	\item The occasional detector failure over the same moving object which would result in a new object every time the detector fails and restores its performance later.
	\item Objects that stop for a while, long enough to be treated as background, and are thus identified as new objects by the detector when they move again.
	\item Objects that move away from the foreground can become too small to be detected by HO-PBAS but they can be tracked by GOTURN and be recognized as previous objects if they come back to the foreground, see Fig.~\ref{fig:accuracy01}.
\end{itemize}    

Next, if the new detection is identified as a new object after the checking, a new tracker is initialized (``Init track'', Fig.~\ref{fig:diagram}). Finally, current detections are saved as previous detections for the next frame (``Save detections'', Fig.~\ref{fig:diagram}).

Although there are more robust state-of-the-art techniques to solve the problem of data association between trackers and detections, such as those well ranked at MOTChallenge benchmark~\cite{mot}, sometimes their low speed is a drawback for real time applications~\cite{henschel2017fusion,keuper2016multi}. Owing to the requirements to develop an end-to-end solution over a low-power embedded platform for real time performance, a more straightforward solution is used.

\tikzstyle{decision} = [diamond, draw, fill=white, text width=4.5em, text badly centered, node distance=3cm, inner sep=0pt]
\tikzstyle{block} = [rectangle, draw, fill=white, text width=5em, text centered, rounded corners, minimum height=4em, node distance=3cm]
\tikzstyle{line} = [draw, -latex']

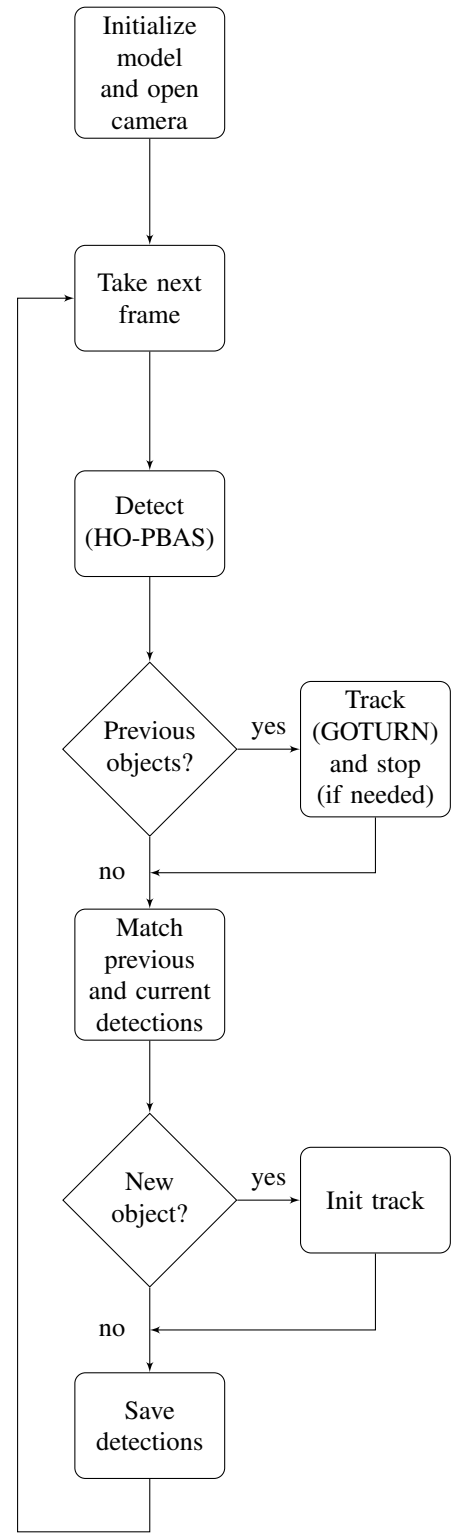
\begin{figure}[h!] \centering    
\begin{tikzpicture}[auto] 
    \node [block] (init) {Initialize model and open camera};
    \node [block, below of=init] (load) {Take next frame};
    \node [block, below of=load] (detect) {Detect (HO-PBAS)};
    \node [decision, below of=detect] (ifprevious) {Previous objects?};
    \node [block, right of=ifprevious, node distance=3cm] (track) {Track (GOTURN) and stop (if needed)};
    \node [block, below of=ifprevious] (matchdetections) {Match previous and current detections};
    \node [decision, below of=matchdetections] (ifnewobject) {New object?};
    \node [block, right of=ifnewobject, node distance=3cm] (inittrack) {Init track};
    \node [block, below of=ifnewobject] (save) {Save detections};
    \path [line] (init) -- (load);
    \path [line] (load) -- (detect);
    \path [line] (detect) -- (ifprevious);
    \path [line] (ifprevious) -- node {yes}(track);
    \path [line] (ifprevious) -- coordinate[pos=0.5](m2)(matchdetections);
    \node [left of=m2, node distance=0.5cm] {no};
    \path [line] (track.south) |-+(0,-1.5em)|- (m2.east);
    \path [line] (matchdetections) -- (ifnewobject);
    \path [line] (ifnewobject) -- node {yes}(inittrack);
    \path [line] (ifnewobject) -- coordinate[pos=0.5](m3)(save);
    \node [left of=m3, node distance=0.5cm] {no};
    \path [line] (inittrack.south) |-+(0,-1.5em)|- (m3.east);
    \path [line] (save.south) |-+(0,-2em)-|+(-5em,0)|-(load.west);
\end{tikzpicture}   
	\caption{Conceptual diagram of the multiple object tracking algorithm proposed in this work.}
	\label{fig:diagram} 
\end{figure}

\section{Results} \label{sec:results}

The proposed algorithm was written in C\texttt{++}, QVGA resolution frames (320$\times$240 pixels) are processed and the maximum RAM usage is 1.6~GB. Following recent trends and suggestions of other authors, which claim a jointly software and hardware design, the results of the present work focus on the frame rate and power consumption of the algorithm, which are hardware dependent~\cite{Sze2016,Sze2017}. 

\subsection{dETRUSC database} \label{sec:accuracy}

The accuracy of a model should be measured on widespread used datasets. The HO-PBAS detector was evaluated on the 2014 updated version of changedetection.net (CDnet) public dataset~\cite{goyette2012changedetection} whereas the GOTURN tracker was tested on VOT2014 dataset~\cite{vot2014}. However, in this work the detector and tracking algorithms were integrated and thus no benchmark for joint performance is available. Additionally, videos are captured with the camera of the Jetson TX2 development kit in order to demonstrate live processing capability for real time decision making. That is, the complete end-node performance aims to demonstrate the feasibility of deep learning techniques for visual tasks on embedded solutions for IoT end-nodes.	

\begin{figure}[!t]
\centering
	\subfloat[Low light intensity and high contrast conditions.]{\includegraphics[width=\columnwidth]{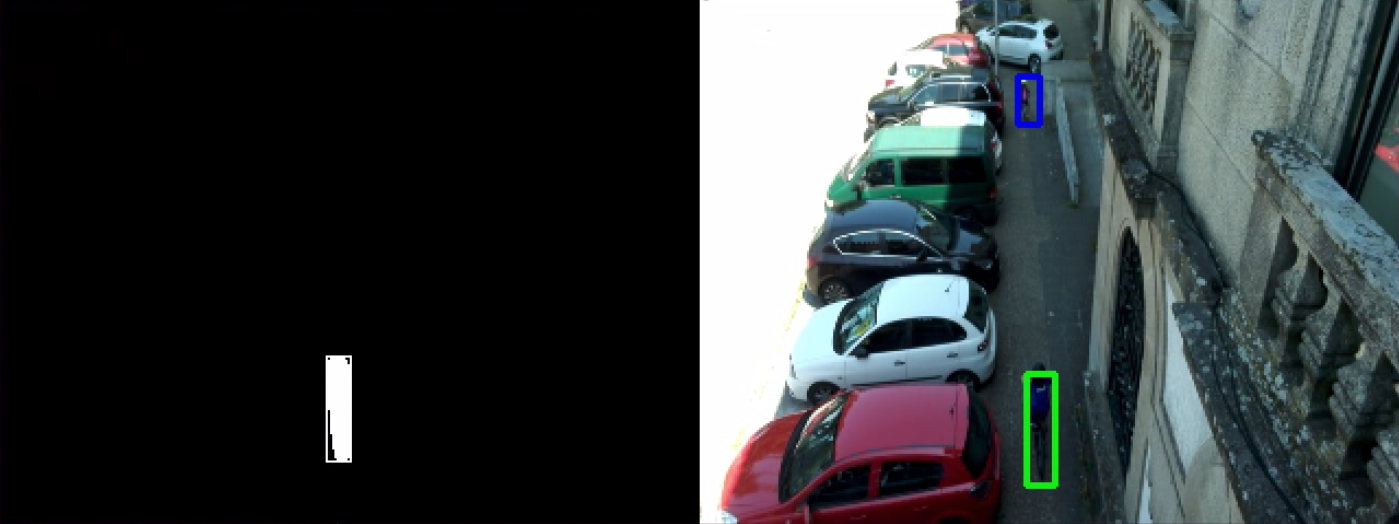}%
	\label{fig:accuracy01}}
	\hfil
	\subfloat[High velocity.]{\includegraphics[width=\columnwidth]{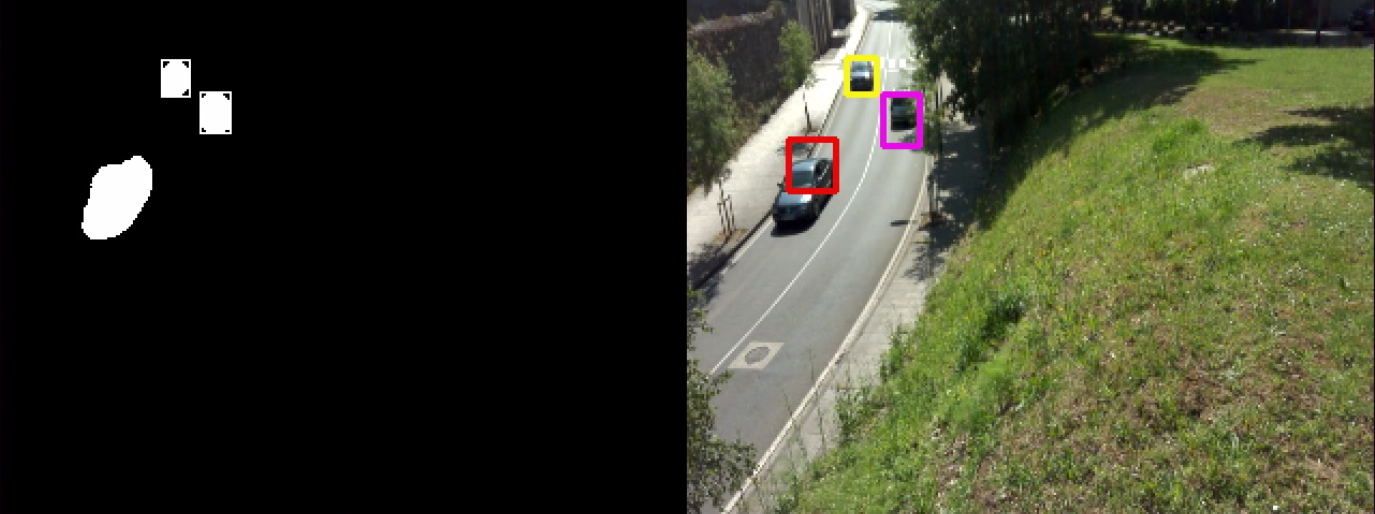}%
	\label{fig:accuracy02}}
\caption{Example of multiple object tracking under different challenging conditions.}
\label{fig:accuracy}
\end{figure}

Due to the aforementioned reasons, a collection of representative sequences captured with the on-board camera were taken and used to exemplify the performance of the proposed algorithm. The dETRUSC video dataset and the results are publicly available and can be accessed at~\cite{citius}. The videos represent real scenarios, including diverse challenging situations such as low light and high contrast conditions, glass reflections, high velocity and shadows. Fig.~\ref{fig:accuracy} shows a pair of captures of the multiple object tracking performance. The black and white image on the left is the HO-PBAS segmentation whereas tracking is depicted on the right. Fig.~\ref{fig:accuracy01} exemplifies the correct tracker performance under low light intensity and high contrast conditions. In Fig.~\ref{fig:accuracy02}, the capability of the tracker to follow vehicles is illustrated, although GOTURN does not handle large movements in order to not increase the complexity of the network and thus achieve high frame rate. 

\subsection{Frame rate}

VOT2017 challenge introduced a real time experiment~\cite{vot2017} but, as explained in the report, tracking speed depends not only on the programming effort and skill but also on the used hardware. If no power constraint nor small size are needed, the latency reduces as a more powerful hardware is chosen. Thus, comparison of different algorithms in terms of throughput (GOPS), latency (s) or frame rate (fps) takes on meaning only when they all run on the same hardware. 

To the best of our knowledge, there are not previous solutions for multiple object detection and tracking running on Jetson TX2, thus this proposal can not be compared with others in terms of velocity. Just in a recent work~\cite{Hanhirova2018}, the latency and throughput of different CNNs only for object recognition or detection are characterized, achieving a maximum frame rate of 4.3~fps in full clocks mode for object detection, what it is not enough for real time inference at the edge. 

\begin{figure}[!t]
\centering
	\subfloat[Normal operation.]{
	
	\begin{tikzpicture}
		\begin{axis}[
			width=\columnwidth,
			height=0.7\columnwidth,
	    	xlabel=\footnotesize{Number of tracked objects},
	    	ylabel=\footnotesize{Frame rate (fps)},
	    	ymin=0,
	    	ymax=17,
	    	legend style={draw=none},
	    	every axis/.append style={font=\footnotesize}]
		    \addplot[smooth,mark=o,orange,very thick,dashed,mark options={solid},mark size=2.5pt] plot coordinates {(1,10.9) (2,8.6) (3,6.8) (4,5.4) (5,4.7) (6,3.9)};
		    \addlegendentry{\footnotesize{Max-N}}
		    \addplot[smooth,mark=*,gold,very thick,dashed,mark options={solid},mark size=2.5pt] plot coordinates {(1,7.2) (2,5.1) (3,4.8) (4,3.9) (5,3.3) (6,2.9)};
		    \addlegendentry{\footnotesize{Max-Q}}
		    \addplot[smooth,mark=square,blue citius,very thick,dashed,mark options={solid},mark size=2.5pt] plot coordinates {(1,8.6) (2,6.5) (3,5.5) (4,4.5) (5,3.7) (6,3.2)};
		    \addlegendentry{\footnotesize{Max-P Core-All}}
		    \addplot[smooth,mark=triangle,green citius,very thick,dashed,mark options={solid},mark size=2.5pt] plot coordinates {(1,10.4) (2,8.1) (3,6.2) (4,5.3) (5,4.5) (6,3.7)};
		    \addlegendentry{\footnotesize{Max-P ARM}}
		    \addplot[smooth,mark=x,royalpurple,very thick,dashed,mark options={solid},mark size=2.5pt] plot coordinates {(1,6.7) (2,5.6) (3,4.7) (4,4.3) (5,3.8) (6,3.4)};
		    \addlegendentry{\footnotesize{Max-P Denver}}
	    \end{axis}
	\end{tikzpicture}
	
	\label{framerate01}}
	\hfil
	\subfloat[Full clocks mode.]{
	
	\begin{tikzpicture}
		\begin{axis}[
			width=\columnwidth,
			height=0.7\columnwidth,
    		xlabel=\footnotesize{Number of tracked objects},
		    ylabel=\footnotesize{Frame rate (fps)},
	    	ymin=0,
	    	ymax=17,
	    	legend style={draw=none},
	    	every axis/.append style={font=\footnotesize}]
	    	\addplot[smooth,mark=o,apricot,very thick,dashed,mark options={solid},mark size=2.5pt] plot coordinates {(1,16.2) (2,9.5) (3,7.4) (4,6.1) (5,5.2) (6,4.5)};
		    \addlegendentry{\footnotesize{Max-N}}
		    \addplot[smooth,mark=*,yellow,very thick,dashed,mark options={solid},mark size=2.5pt] plot coordinates {(1,9.3) (2,6.7) (3,5.3) (4,4.3) (5,3.7) (6,3.2)};
		    \addlegendentry{\footnotesize{Max-Q}}
		    \addplot[smooth,mark=square,processcyan,very thick,dashed,mark options={solid},mark size=2.5pt] plot coordinates {(1,13.1) (2,8.2) (3,6.4) (4,5.2) (5,4.4) (6,3.9)};
		    \addlegendentry{\footnotesize{Max-P Core-All}}
		    \addplot[smooth,mark=triangle,springgreen,very thick,dashed,mark options={solid},mark size=2.5pt] plot coordinates {(1,12.5) (2,8.8) (3,6.8) (4,5.5) (5,4.7) (6,4.1)};
		    \addlegendentry{\footnotesize{Max-P ARM}}
		    \addplot[smooth,mark=x,orchid,very thick,dashed,mark options={solid},mark size=2.5pt] plot coordinates {(1,9.0) (2,7.2) (3,5.8) (4,5.1) (5,4.2) (6,3.8)};
		    \addlegendentry{\footnotesize{Max-P Denver}}
		\end{axis}
	\end{tikzpicture}
	
	\label{framerate02}}
	
\caption{Frame rate of the multiple object tracking algorithm for different number of tracked objects and Jetson TX2 operation modes.}
\label{fig:framerate}
\end{figure}
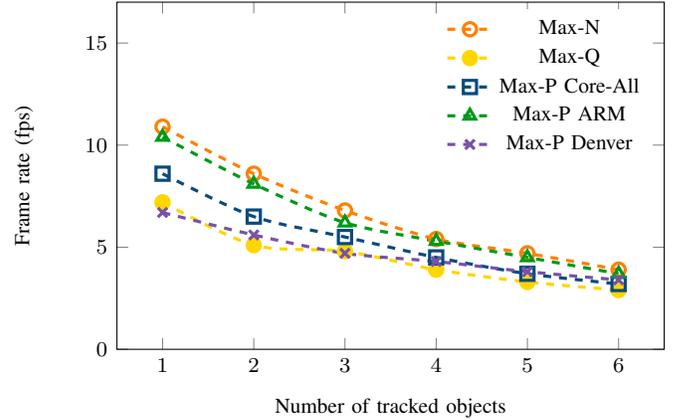
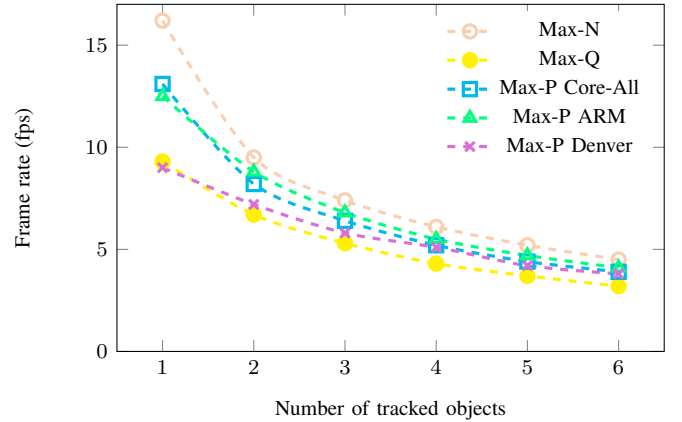

In order to measure the maximum algorithm speed over the Jetson, regardless of the capture frame rate, prerecorded video processing is performed instead of live capture. The metrics for the frame rate of the proposed model as a function of the number of tracked objects were measured for all the available operation modes and full clocks mode, Fig.~\ref{fig:framerate}. As the number of objects increases, the algorithm processes the frames more slowly, although a saturation tendency is observed. The most efficient operation mode, Max-Q, also offers the worst performance in terms of velocity. Regarding the full clocks mode, it improves the one object tracking performance but no significant profit is observed for multiple object purpose.

In view of these results, real time performance was evaluated at Max-N operation mode and 10~fps video capture, using at any time only the available frame from the live camera without storing intermediate frames for delayed processing. Satisfactory results were found in outdoor and indoor scenarios, under low light intensity and high contrast conditions and fast moving objects such as vehicles. Video results can be accessed at the dETRUSC database site~\cite{citius}.

\subsection{Power consumption}

\begin{figure*} \centering
\begin{tikzpicture}
	\begin{axis}[
		width=\textwidth,
	    height=0.3\textwidth,
	    ybar stacked,
	    bar width=0.3cm,
	    xlabel=\footnotesize{Number of tracked objects},
	    ylabel=\footnotesize{P (W)},
	    symbolic x coords={1,2,3,4,5,6},
	    xtick=data,
	    ymin=0,ymax=15,
	    bar shift=-0.8cm,
	    legend style={at={(0.27,0.95)},anchor=north,legend columns=-1,name = serieA,draw=none},
	    every axis/.append style={font=\footnotesize}
	    ]
		\addplot[fill=orange] coordinates {(1,7.94) (2,9.27) (3,9.76) (4,9.76) (5,10.18) (6,9.93)};
		\addplot[fill=apricot] coordinates {(1,2.86) (2,1.69) (3,1.68) (4,2.03) (5,1.74) (6,2.04)};
		\legend{\footnotesize{Max-N}}
	\end{axis}
	\begin{axis}[
		width=\textwidth,
	    height=0.3\textwidth,
	    ybar stacked,
	    bar width=0.3cm,
	    xtick=data,
	    ymin=0,ymax=15,
	    bar shift=-0.4cm,
	    legend style={at={([xshift = -2mm]serieA.north)},anchor=north,legend columns=-1,name = serieB,draw=none},
	    every axis/.append style={font=\footnotesize}
	    ]
		\addplot[fill=gold] coordinates {(1,4.57) (2,5.01) (3,5.71) (4,5.92) (5,6.15) (6,6.21)};
		\addplot[fill=yellow] coordinates {(1,0.39) (2,0.65) (3,0.30) (4,0.26) (5,0.22) (6,0.28)};
		\legend{\footnotesize{Max-Q}}
	\end{axis}
	\begin{axis}[
		width=\textwidth,
	    height=0.3\textwidth,
	    ybar stacked,
	    bar width=0.3cm,
	    xtick=data,
	    ymin=0,ymax=15,
	    legend style={at={([xshift = 3.5mm]serieB.north)},anchor=north,legend columns=-1,name = serieC,draw=none},
	    every axis/.append style={font=\footnotesize}
	    ]
		\addplot[fill=blue citius] coordinates {(1,5.67) (2,6.36) (3,6.91) (4,7.24) (5,7.29) (6,7.33)};
		\addplot[fill=processcyan] coordinates {(1,1.80) (2,1.45) (3,1.36) (4,1.26) (5,1.41) (6,1.48)};
		\legend{\footnotesize{Max-P Core-All}}
	\end{axis}
	\begin{axis}[
		width=\textwidth,
	    height=0.3\textwidth,
	    ybar stacked,
	    bar width=0.3cm,
	    xtick=data,
	    ymin=0,ymax=15,
	    bar shift=0.4cm,
	    legend style={at={([xshift = 7mm]serieC.north)},anchor=north,legend columns=-1,name = serieD,draw=none},
	    every axis/.append style={font=\footnotesize}
	    ]
		\addplot[fill=green citius] coordinates {(1,6.44) (2,7.52) (3,7.79) (4,8.36) (5,8.54) (6,8.23)};
		\addplot[fill=springgreen] coordinates {(1,1.02) (2,0.80) (3,0.91) (4,0.60) (5,0.68) (6,1.06)};
		\legend{\footnotesize{Max-P ARM}}
	\end{axis}
	\begin{axis}[
		width=\textwidth,
	    height=0.3\textwidth,
	    ybar stacked,
	    bar width=0.3cm,
	    xtick=data,
	    ymin=0,ymax=15,
	    bar shift=0.8cm,
	    legend style={at={([xshift = 6mm]serieD.north)},anchor=north,legend columns=-1,draw=none},
	    every axis/.append style={font=\footnotesize}
	    ]
		\addplot[fill=royalpurple] coordinates {(1,5.20) (2,5.95) (3,6.42) (4,7.05) (5,7.51) (6,7.87)};
		\addplot[fill=orchid] coordinates {(1,0.91) (2,1.25) (3,1.38) (4,1.33) (5,1.04) (6,1.03)};
		\legend{\footnotesize{Max-P Denver}}
	\end{axis}
\end{tikzpicture}
	\caption{Total power consumption of the system for different number of tracked objects and Jetson TX2 operation modes. The bar increment represents the additional energy consumption on full clocks mode.}
	\label{fig:totalpower-clocks} 
\end{figure*}
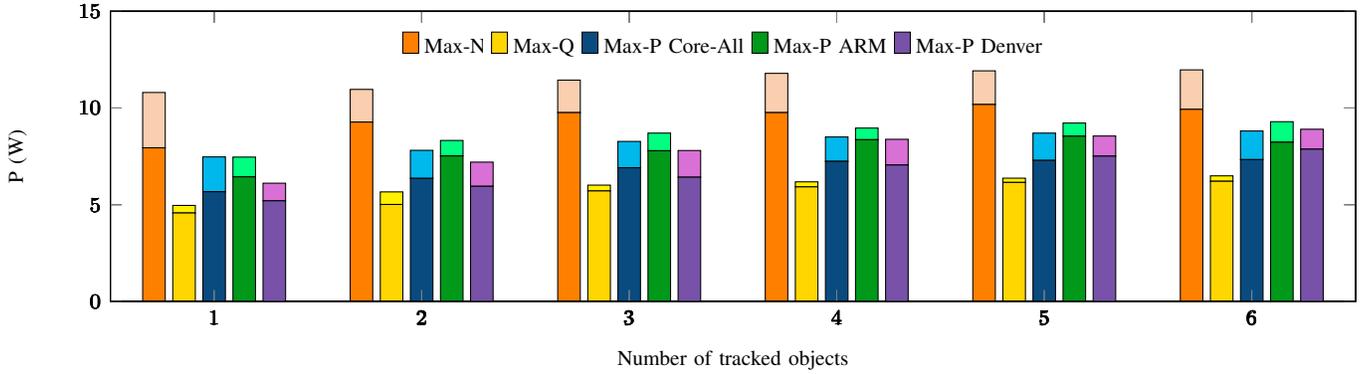

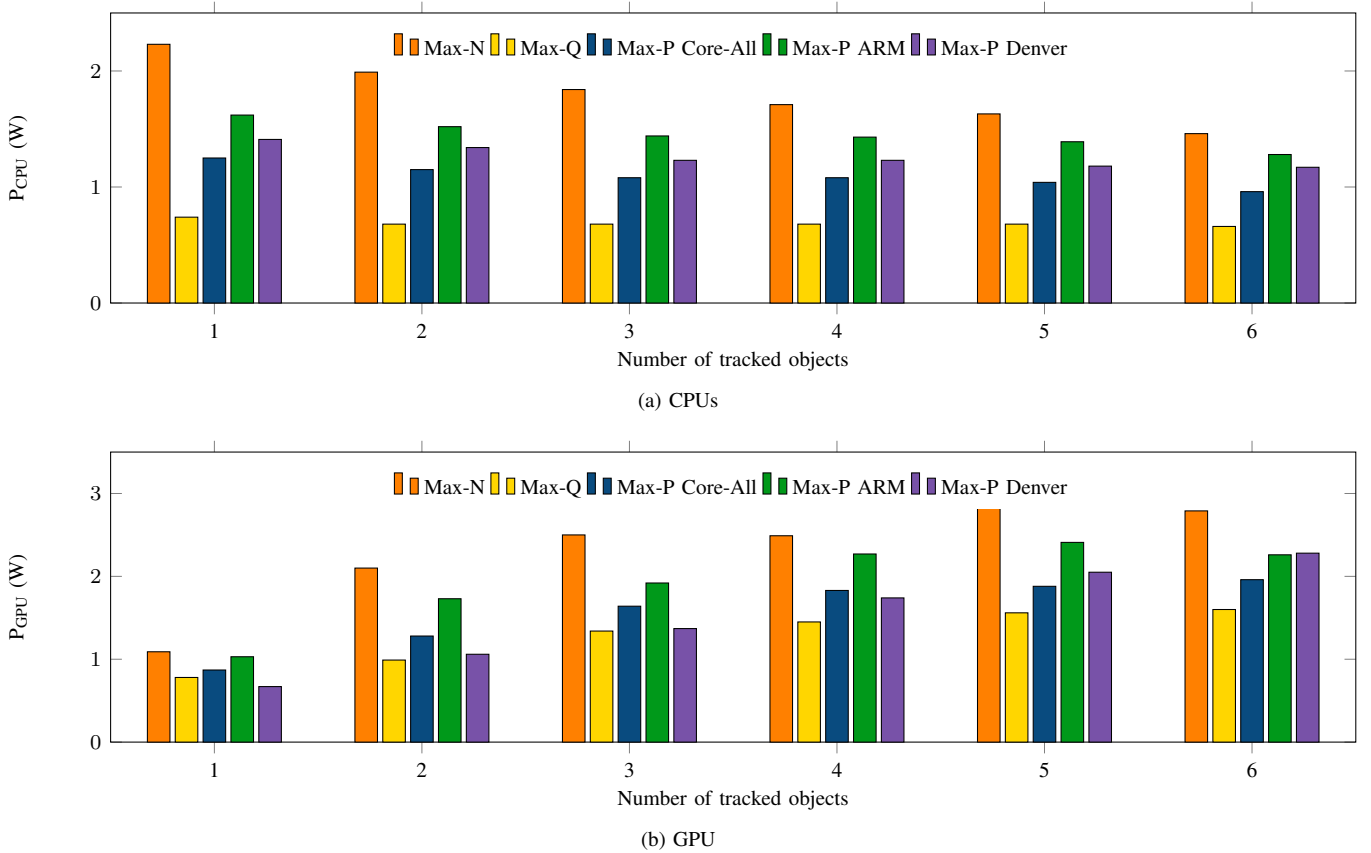
\begin{figure*}[!t]
\centering
\subfloat[CPUs]{
	\begin{tikzpicture}
		\begin{axis}[
			width=\textwidth,
	    	height=0.3\textwidth,
		    ybar,
		    bar width=0.3cm,
		    xlabel=\footnotesize{Number of tracked objects},
		    ylabel=\footnotesize{{P$_\textrm{CPU}$ (W)}},
		    symbolic x coords={1,2,3,4,5,6},
		    xtick=data,
		    ymin=0,ymax=2.5,
		    legend style={at={(0.5,0.95)},anchor=north,legend columns=-1,draw=none},
	    	every axis/.append style={font=\footnotesize}
		    ]
			\addplot[fill=orange] coordinates {(1,2.23) (2,1.99) (3,1.84) (4,1.71) (5,1.63) (6,1.46)};
			\addplot[fill=gold] coordinates {(1,0.74) (2,0.68) (3,0.68) (4,0.68) (5,0.68) (6,0.66)};
			\addplot[fill=blue citius] coordinates {(1,1.25) (2,1.15) (3,1.08) (4,1.08) (5,1.04) (6,0.96)};
			\addplot[fill=green citius] coordinates {(1,1.62) (2,1.52) (3,1.44) (4,1.43) (5,1.39) (6,1.28)};
			\addplot[fill=royalpurple] coordinates {(1,1.41) (2,1.34) (3,1.23) (4,1.23) (5,1.18) (6,1.17)};
			\legend{\footnotesize{Max-N},\footnotesize{Max-Q},\footnotesize{Max-P Core-All},\footnotesize{Max-P ARM},\footnotesize{Max-P Denver}}
		\end{axis}
	\end{tikzpicture}
	
	\label{fig:power-cpu}}
	\hfil
	\subfloat[GPU]{

	\begin{tikzpicture}
		\begin{axis}[
			width=\textwidth,
	    	height=0.3\textwidth,
		    ybar,
		    bar width=0.3cm,
		    xlabel=\footnotesize{Number of tracked objects},
		    ylabel=\footnotesize{{P$_\textrm{GPU}$ (W)}},
		    symbolic x coords={1,2,3,4,5,6},
		    xtick=data,
		    ymin=0,ymax=3.5,
		    legend style={at={(0.5,0.95)},anchor=north,legend columns=-1,draw=none},
	    	every axis/.append style={font=\footnotesize}
		    ]
			\addplot[fill=orange] coordinates {(1,1.09) (2,2.10) (3,2.50) (4,2.49) (5,2.86) (6,2.79)};
			\addplot[fill=gold] coordinates {(1,0.78) (2,0.99) (3,1.34) (4,1.45) (5,1.56) (6,1.60)};
			\addplot[fill=blue citius] coordinates {(1,0.87) (2,1.28) (3,1.64) (4,1.83) (5,1.88) (6,1.96)};
			\addplot[fill=green citius] coordinates {(1,1.03) (2,1.73) (3,1.92) (4,2.27) (5,2.41) (6,2.26)};
			\addplot[fill=royalpurple] coordinates {(1,0.67) (2,1.06) (3,1.37) (4,1.74) (5,2.05) (6,2.28)};
			\legend{\footnotesize{Max-N},\footnotesize{Max-Q},\footnotesize{Max-P Core-All},\footnotesize{Max-P ARM},\footnotesize{Max-P Denver}}
		\end{axis}
	\end{tikzpicture}

	\label{fig:power-gpu}}
	\caption{Power consumption of the CPUs and GPU for different number of tracked objects and Jetson TX2 operation modes.}
	\label{fig:power-cpu-gpu}
\end{figure*}

Consumption is one of the main limiting factors for deep learning application on embedded IoT end-nodes. The opportunities that CNNs offer for image processing are undeniable but their practical use and expansion will highly depend on the hardware design and the development of hardware-oriented algorithms~\cite{Sze2016}. The energy consumption of CNNs is dominated by data movement instead of the computation itself~\cite{Sze2017}. Fortunately, the most costly operations in terms of data movement are highly parallel. 

The total power consumption of the proposed multiple object tracking algorithm running on NVIDIA Jetson TX2 development kit is shown in Figure~\ref{fig:totalpower-clocks}. It was measured for the different operation modes of the board summarized in Table~\ref{tx2modes}. The increments of the bars represent the additional consumption due to full clocks mode activation, which forces to run all cores at the maximum speed all time. As can be observed, the latter has little effect for Max-Q mode since it limits the clocks to ensure operation in the most efficient range. The power was also measured as a function of the number of tracked objects, finding that power consumption softly increases with the presence of more moving objects in the scene. An analogous representation of the power consumed by the CPUs (P$_\textrm{CPU}$) and GPU (P$_\textrm{GPU}$) is depicted in Fig.~\ref{fig:power-cpu-gpu}. An increment of the GPU utilization is observed through a greater power consumption as the number of tracked objects increases. On the contrary, CPUs consumption diminishes as GPU workload raises because they must await GPU completion.

As expected, the Max-Q operation mode is the most cost-effective in terms of energy, regardless of metric, ranging from 4.57 to 6.21~W. On the contrary, Max-N is the most expensive with 12~W maximum consumption since all CPUs and GPU run at maximum clock speeds, see Table~\ref{tx2modes}. Max-P modes are a balance between both, running all CPUs, only ARM A57 processors, or only Denver processors. That is to say, using a 3S 6000~mAh LiPo battery as in our experimental set-up, an autonomy of six hours under continuous performance at maximum power consumption is guaranteed. 

If we compare our proposal with the state-of-the-art detectors and trackers we can observe that improving the accuracy has a great impact on velocity and power consumption. For instance, Faster R-CNN~\cite{ren2017faster} and R-FCN~\cite{dai2016r} detectors can run at 5-17~fps and 6~fps, respectively, on a K40 GPU, which has a power consumption of 235~W. That is, a similar frame rate as ours is achieved but performing only the detection and with 20 times more power consumption. In turn, MDNet~\cite{nam2016learning} tracker achieves a speed of 1~fps running on a K20 GPU, 225~W power consumption, whereas high frame rate of 58-86~fps is obtained with SiameseFC~\cite{bertinetto2016fully} tracker but using a GeForce GTX Titan X, which consumes 250~W. Moreover, those data correspond to only 1-object tracking and do not consider the detection phase. Additionally, we are not considering CPUs consumption in these detectors and trackers results, nor comparing dimensions and price of the hardware, nor memory usage, which are also penalized. Summarizing, what we propose is an algorithm that solves the application problem in real time, with low memory usage, and implemented on a small dimensions hardware platform, which also has an affordable price and with a low-power consumption of 12~W.

\section{Conclusions} \label{sec:conclusions}
An end-to-end solution for real time deep learning-based multiple object tracking in an embedded and low-power IoT oriented platform is presented. NVIDIA Jetson TX2~\cite{jetsontx2} development kit was chosen because it allows development from beginning to end from the early stage, including camera and wireless connection. It was powered by a LiPo battery and remotely controlled using a tablet. Another strengths of the platform are price and dimensions.

The Hardware-Oriented PBAS (HO-PBAS) algorithm~\cite{garcia2017effect} to detect moving objects was integrated with the GOTURN CNN based tracker~\cite{Held2016} in order to perform multiple object tracking. Whereas both algorithms present good results separately, additional development was needed to overcome some limitations due to the jointly operation. 

Qualitative results over the dETRUSC video dataset captured with the on-board camera can be found at~\cite{citius}. Good results were found under real challenging scenarios including low light and high contrast conditions, glass reflections, high velocity and shadows. Regarding velocity and power consumption, real time performance at 10~fps video capture, using at any time only the available frame from the live camera without intermediate storage for delayed processing, with a total power consumption of only 12~W is achieved.

\bibliographystyle{IEEEtran}
\bibliography{IEEEabrv,biblio}

\vfill


\end{document}